\documentclass[10pt,twocolumn,letterpaper]{article}

\usepackage{cvpr}
\usepackage{times}
\usepackage{epsfig}
\usepackage{graphicx}
\usepackage{amsmath}
\usepackage{amssymb}
\usepackage{multirow}
\usepackage{subfig}
\usepackage{url}
\usepackage{multirow}
\usepackage{breqn}
\usepackage{makecell}
\usepackage{arydshln}

\usepackage[pagebackref=true,breaklinks=true,letterpaper=true,colorlinks,bookmarks=false]{hyperref}
\usepackage{cleveref}
 \cvprfinalcopy % *** Uncomment this line for the final submission

\def\cvprPaperID{7439} % *** Enter the CVPR Paper ID here

\ifcvprfinal\fi
\begin{document}

%%%%%%%%% TITLE
\title{Mixture Dense Regression for Object Detection and Human Pose Estimation}

\author{Ali Varamesh, Tinne Tuytelaars\\
ESAT-PSI, KU Leuven\\
%Center for Processing Speech and Images, ESAT, KU Leuven\\
% \texttt{\{ali.varamesh,tinne.tuytelaars\}@kuleuven.be} \\
{\tt\small \{ali.varamesh,tinne.tuytelaars\}@esat.kuleuven.be}
}

% \maketitle

\twocolumn[{%
\renewcommand\twocolumn[1][]{#1}%
\maketitle
% \begin{center}
    % \centering
    % \includegraphics[width=.9\textwidth]{images/comp_area_vis.jpg}
    % \captionof{figure}{Distribution of scale of objects predicted by components of mixture models compared to the ground truth object scale distribution on \textit{coco-val} set. $MDN_{x}$ denotes mixture model with x components and vertical axis indicates scale. Mixture components are named with respect to scale of their predictions. The number bellow each box is the portion of instances of each scale in the dataset or portion of predictions made by a mixture component}
    
    % % For the ground truth box plot, bellow each box is the portion of instances of each scale in the dataset. For the mixture models plots, the number bellow each box is the .}
    %  \label{fig:scale_distribution}
% \end{center}%
}]

\thispagestyle{empty}

%%%%%%%%% ABSTRACT
\begin{abstract}

% Mixture models are well-established machine learning approaches that, in computer vision, have mostly been applied to inverse or ill-defined problems. 
Mixture models are well-established learning approaches that, in computer vision, have mostly been applied to inverse or ill-defined problems. However, they are general-purpose divide-and-conquer techniques, splitting the input space into relatively homogeneous subsets in a data-driven manner. Not only ill-defined but also well-defined complex problems should benefit from them. To this end, we devise a framework for spatial regression using mixture density networks. We realize the framework for object detection and human pose estimation. For both tasks, a mixture model yields higher accuracy and divides the input space into interpretable modes. For object detection, mixture components focus on object scale, with the distribution of components closely following that of ground truth the object scale. This practically alleviates the need for multi-scale testing, providing a superior speed-accuracy trade-off. For human pose estimation, a mixture model divides the data based on viewpoint and uncertainty -- namely, front and back views, with back view imposing higher uncertainty. We conduct experiments on the MS COCO dataset and do not face any mode collapse.  
% of multi-modal spatial outputs
% However, to avoid numerical instabilities, we had to modify the activation function for the mixture variance terms slightly.

%which allows achieving higher accuracy and gaining least control over a model's internal logic. 
%on dense object detection and human pose estimation 
% Unlike previous works, we show that by carefully initializing the kernel parameters, a mixture model in fact can be successfully trained at a large scale without facing mode collapse. Our findings indicate importance of deploying principled multi modal techniques in deep convolutional networks which operate on highly complex and multi-modal data. From another perspective, we illustrate how interpretability in model predictions can be achieved by a multi-modal design.
\end{abstract}

%%%%%%%%% BODY TEXT
\vspace{-5pt}
\section{Introduction}

% Dense prediction tasks on RGB images, including object detection, instance segmentation, and human pose estimation, have seen revolutionary gains in accuracy thanks to the fast advancements in deep convolutional neural networks in the recent years. 
Over the span of a few years, there has been massive progress in designing increasingly efficient architectures, loss functions, and optimization procedures for mainstream computer vision tasks such as image classification, object detection, semantic segmentation, and pose estimation~\cite{krizhevsky2012imagenet,simonyan2014very,szegedy2015going,newell2016stacked,lin2017focal,kingma2014adam,berman2018lovasz,he2016deep,ren2015faster}. However, from a machine learning perspective, there is still a lot to be desired. For example, when it comes to capturing the multi-modal nature of visual data, most of the solutions for object detection leave it to the optimizer to figure it all out. Still, given the fundamental limitations of machine learning \cite{wolpert1997no,wolpert1996lack}, this is an unrealistic expectation. Modeling a multi-modal distribution using a single-mode model will always lead to sub-optimal predictions.

%we are still far from the state of having best solutions from machine learning viewpoint. 

% As a case in point, let us consider dense detection and estimation tasks (e.g.~object detection or pose estimation) 
As a case in point, let us consider dense object detection. For a given input image, at each spatial location, a model should have a classification output and do a spatial regression. The classification component naturally is a multi-modal problem. As such, any solution has to learn different modalities, commonly realized via multinomial classification. For the spatial regression, on the other hand, either there are no such discernible modalities, or it is not straightforward how to model them. In object detection, it could be the categories that govern the bounding box regression, mere foreground vs. background segmentation, or a coarser categorization. We cannot say for sure which one it is. Similarly, in dense human pose estimation using offset regression, there is a separate output for each body part. But, given a part, either the scale or the pose of a person could be the dominant mode for the regression task.

\begin{figure}[t]
\centering
\resizebox{.5\textwidth}{!}{
%\framebox[4.0in]{$\;$}
\includegraphics[width=.7\textwidth]{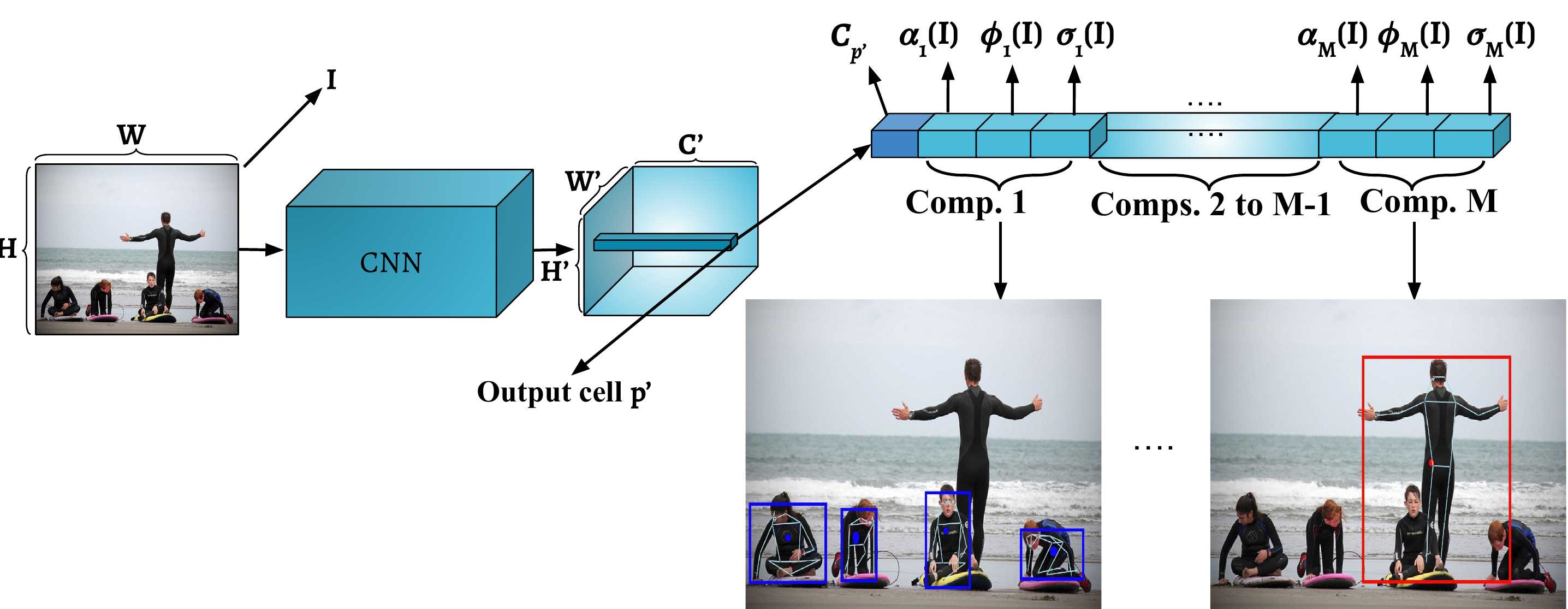}
}
% \caption{The proposed mixture spatial regression framework.}
\caption{The proposed mixture spatial regression framework. When realized for human pose estimation, two modes are retrieved based on viewpoint.}
% \caption{The proposed mixture spatial regression framework. For human pose estimation, two modes are retrieved for dealing with viewpoint.}
% The proposed mixture spatial regression framework. For human pose estimation, two modes are retrieved from the data, specializing in instances with (or otherwise closer to) front view and back view, respectively.
\label{fig:schematic}
\end{figure}

As one can see, explicitly identifying the underlying modes is often impossible. However, in machine learning, there are well-established techniques for dealing with multi-modality. For example, mixture models \cite{mclachlan1988mixture}, including mixture density networks \cite{bishop1994mixture} and mixture of experts \cite{jordan1994hierarchical},
% or CARTs \cite{breiman2017classification} 
are powerful techniques for imposing structure on a model's prediction mechanism,  which leads to higher accuracy and model interpretability. In particular, they divide the input space based on the relationship between input and output. In other words, depending on the targets, they optimally split the input space so that higher performance can be achieved. Of course, these techniques are not unknown to our community, but so far, they have been almost exclusively applied to ill-defined problems like 3D pose estimation \cite{li2019generating, ye2018occlusion} or video frame prediction~\cite{makansi2019overcoming}.

%it seems impossible to find a certain answer for such questions. 
%such that it helps with gaining \cvprPaperID{}
%strange 

In this work, we advocate a more extensive use of mixture models. 
We show how to improve bottom-up dense object detection and human pose estimation tasks by incorporating their spatial regression sub-tasks in a mixture density network. Our framework yields significant improvements for both tasks in terms of accuracy, speed-accuracy trade-off, convergence, and interpretability. To the best of our knowledge, 
we are the first to successfully integrate mixture density networks into 2D object detection and human pose estimation tasks. We have published the source-code\footnote{https://github.com/alivaramesh/MixtureDenseRegression}. 
 The following is a summary of our contributions:
\begin{itemize}
    
\item To account for the multi-modal nature of the visual domain, we propose a new formulation for dense spatial localization using mixture density networks, which proves to be superior to single-mode models.

\item 
%We conduct extensive experiments and illustrate 
We show that a mixture object detection model learns to deal with object scale variation through different components, provides significantly better speed-accuracy trade-off, and converges faster.

% leads to significantly faster convergence.

%. More importantly, we show that a mixture solution 
\item We model offset regression in human pose estimation using a mixture model. Our model yields significant gain in accuracy and reveals viewpoint as the dominant multi-modal factor. Further analysis shows that, in fact, uncertainty decides mixture components.
%that 
\end{itemize}

\section{Related work}

%\subsection{Object detection and human pose estimation}

% Lets start with the precise definition of the tasks we consider in this work. For object detection, the goal is to find all instance of all categories of interest in a given image and generate a tight bounding box around the each object. For human pose estimation, the goal is to localize a predefined set of keypoints for each person and have the keypoints grouped together. 

% Modern solutions for object detection and human pose estimation either use a top-down design or a bottom-up one. This work focuses on single-stage bottom-up models. Here, we review related work from both approaches.

Modern solutions for object detection and human pose estimation either use a top-down design or a bottom-up one. Here, we review related work from both approaches. However, our framework is based on the bottom-up approach.

\subsection{Top-down models}
In top-down models, an image is first processed by a CNN to propose an initial set of regions that may include objects of interest. The proposals are then further processed for more accurate recognition and localization. For object detection, this means classification of the object in a given region and generating a tight bounding box around it \cite{girshick2014rich,ren2015faster,he2017mask,singh2018sniper}. Before the second stage, the regions are resized to a fixed size, therefore gaining some built-in robustness to scale variation. In top-down human pose estimation, first, a set of regions containing persons are generated by an object detector; next within each region, a predefined set of body keypoints (e.g. eyes, shoulders, etc.) are localized \cite{newell2016stacked,li2019rethinking,he2017mask,chen2018cascaded,li2019rethinking}. In general, two-stage procedures are more accurate than single-stage models but incur a significant delay. A state-of-the-art model can may about a second to process an image\cite{li2019scale}. 

% In a sense, these models refine an initial set of proposals.
% The best performing (in terms f AP) model  \cite{li2019scale}).  

\subsection{Bottom-up models}
In a bottom-up approach, in a single stage, a model simultaneously does classification at all given spatial locations in a dense fashion and also estimates pose parameters. Classification head determines if a location indicates the center of an object, or if a location is inside an object region.  In object detection, desired pose parameters represent object bounding boxes \cite{liu2016ssd,redmon2018yolov3,lin2017focal,zhou2019objects}. 

In bottom-up human pose estimation, the traditional approach is to generate a dense heatmap for each body part to predict the presence of that part at each spatial location. Simultaneously, at each location, an embedding is generated to distinguish keypoints of different person instances \cite{cao2017realtime,papandreou2018personlab,newell2017associative}. In the other approach, every location is classified as center of a person or not, and an offset vector is generated from that location to each body part \cite{zhou2019objects}. This method is faster and eliminates the sub-optimal post-processing step for grouping keypoints using embeddings generated by the former approach. However, it is more challenging to optimize. In fact, in \cite{zhou2019objects}, offset regression does not deliver high spatial precision, and body part heatmaps are used to refine the predictions, incurring a delay. A central motivation for our work has been to improve offset regression such that a refinement step becomes unnecessary.

% \vspace{-5pt}
\subsection{Multiple choice models}
Multiple choice models include approaches where, for a given input, a model makes multiple predictions, from which one is chosen as the best. In spirit, they are similar to mixture models. In the context of image classification, it is shown by many works that generating multiple diverse predictions works better than a single head or an ensemble of models \cite{guzman2012multiple, lee2015m,lee2016stochastic, rupprecht2017learning}. However, they depend on an oracle to choose the best prediction for a given input. This may be fine when another downstream application further examines the predictions, but it is a big obstacle for the widespread deployment of such models. Additionally, unlike mixture density networks, these methods do not have a mechanism to learn the density of outputs conditioned on the input.

Mixture density networks \cite{bishop1994mixture} have attracted a lot of attention in recent years. In particular, it has been applied to 3D human pose estimation \cite{li2019generating} and 3D hand pose estimation \cite{ye2018occlusion}. In 2D human pose estimation, \cite{rupprecht2017learning} have reported an unsuccessful application of MDNs, failed due to numerical instabilities.  Here, we show that properly modifying the variance activation function for Gaussian kernels eliminates such instabilities. MDNs are also used by Prokudin et al. \cite{prokudin2018deep} for quantifying uncertainty in angular pose estimation. Nevertheless, to the best of our knowledge, mixture density networks have not been adapted for mainstream vision tasks of object detection and human pose estimation on large scale real-world datasets.

\section{Methodology}

In this section, first, we review the mixture density networks. Next, we illustrate how to model dense spatial regression using a mixture model. 

\subsection{Mixture Density Networks}
Mixture models are powerful tools for estimating the density of any distribution \cite{mclachlan1988mixture}. Ideally, they recover modes that contribute to the generation of data and their distribution. For a regression task, mixture models help to avoid converging to an average target given an input. For example, consider estimating the density of data generated by a bi-modal Gaussian distribution. Using a single Gaussian model will deliver sub-optimal results by predicting a mean value squashed in between two actual centers. However, a mixture model avoids this issue by assigning data points to proper generators. As in the example above, mixture models are straightforward to interpret.

In the context of neural networks, mixture density networks (MDN) \cite{bishop1994mixture} enable us to use a neural network to estimate the parameters of a mixture model. An MDN estimates the probability density of a target vector conditioned on the input. Assume a regression task on a dataset with the set of input vectors denoted by $\displaystyle \{\textbf{x}_0 \dots \textbf{x}_n\}$ and the associated target vectors $\displaystyle \{\textbf{t}_0 \dots \textbf{t}_n\}$. The objective of MDN is to fit the weights of a neural network such that it maximizes the likelihood of the training data. The key issue then is to formulate the probability density of the target conditioned on the input. Eq. \ref{eq:mdn_prob_den} shows how this is done in MDNs.
\vspace{-5pt}

\begin{equation} \label{eq:mdn_prob_den}
p(\textbf{t}_i|\textbf{x}_i) = \sum_{m=1}^{M} \alpha_m(\textbf{x}_i)\phi_m(\textbf{t}_i|\textbf{x}_i)
\end{equation}
% }
\vspace{-5pt}

In Eq. \ref{eq:mdn_prob_den}, M is a hyper-parameter denoting the number of components constituting the mixture model. $\displaystyle\alpha_m(\textbf{x}_i)$ is called mixing coefficient and indicates the probability of component $\displaystyle m$ being responsible for generation of the sample $\displaystyle \textbf{x}_i$. $\displaystyle \phi_m$ is the probability density function of component $\displaystyle m$ for computing density of $\displaystyle \textbf{t}_i$ conditioned on $\displaystyle \textbf{x}_i$. The density function can be chosen from a wide set of well-known kernels. In practice Gaussian kernel (given in Eq. \ref{eq:gaussain_cpdf}) works well and is the most common choice.

\vspace{-3pt}
% {
% \centering
% \resizebox{.5\textwidth}{!} 
% {
 \begin{dmath}  \label{eq:gaussain_cpdf}
% \begin{equation}
\phi_m(\textbf{t}_i|\textbf{x}_i) = \frac{1}{(2\pi)^{c/2}\sigma_m(\textbf{x}_i)^c} exp\Big\{- \frac{||  \textbf{t}_{i} - \boldsymbol{\mu}_{m}(\textbf{x}_i) ||^{2}}   {2\sigma_m(\textbf{x}_i)^2}\Big\} 
\end{dmath}
% \end{equation}
% } 
% }
\vspace{-3pt}

In Eq. \ref{eq:gaussain_cpdf}, $c$ indicates dimension of the target vector, $\boldsymbol{\mu}_{m}$ is the mean of component $m$, and $\sigma_m$ is the common variance parameter for component $m$. The variance term does not have to be shared between dimensions of target space and can be replaced with a diagonal or full covariance matrix if necessary \cite{bishop1994mixture}. Note that MDNs do not presume independence among the components of the target vector $\displaystyle \textbf{t}$. To be more precise, elements of the $\displaystyle \textbf{t}$ are independent given a mixture component; however, the full model enforces dependence among the elements of the target, through learning different modes of the data using each mixture component. 

\subsection{Mixture dense spatial regression}

In this section, we illustrate how to formulate both object detection and human pose estimation tasks using mixture density networks. We develop our formulation on top of the recent CenterNet dense detection model \cite{zhou2019objects}. The general formulation is as follows. 
Given an image, each spatial location needs to be classified to determine whether it is representing the center of an instance. The classification branch is realized by doing dense binary classification for each category $\displaystyle y \in \textbf{Y}$. The number of categories for object detection is equal to the number of classes in a dataset, and for human pose estimation, it only includes the person class. Beside the classification branch, at each location we also need to predict pose parameters of interest $\displaystyle \textbf{T}$ \cite{amit2014object}. For object detection, pose parameters correspond to the height and width of an object, therefore a 2-dimensional vector. For human pose estimation, $\displaystyle \textbf{T}$ includes $\displaystyle \textit{K}$ $2D$ dimensional offset vectors from the center of a person to each of the $\displaystyle \textit{K}$ body parts (K is 17 in the MS COCO keypoints dataset). This formulation is, in particular, efficient for human pose estimation. Unlike the top-down methods, there is no need to use an object detector to localize person instances first. And, unlike traditional bottom-up methods, the grouping of body parts is not left as a post-processing step based on learned embeddings. Rather, at every spatial location, the model predicts if it is the center of a person and generates an offset vector to the location of each keypoint. 

The most common loss for training the spatial pose parameters is the $\displaystyle L_1$ loss function \cite{papandreou2018personlab,kreiss2019pifpaf,cao2017realtime,zhou2019objects}. However, spatial regression is a multi-modal task, and we believe modeling it using a single-mode network will lead to a sub-optimal prediction. Therefore, we use a mixture density network to model the spatial regression task. Now we proceed to describe our mixture dense prediction model formally.

Given an input RGB image  $\displaystyle I$ of size $H*W*3$, a CNN processes $\displaystyle I$ and generates an output of dimensions $H'*W'*C'$. Here we have $\displaystyle H = D*H'$ (and similarly for width), where $\displaystyle D$ is the down sampling factor of the network. We indicate the set of all output cells using $P$. At $p \in P$, the output channels $C'$, includes $Y$ binary classification channels. It also includes pose parameters $T$. For object detection, $T$  is a 2D vector of the form: $\displaystyle T = [p_w,p_h]$ corresponding to width and height of an object. For human pose estimation, $T$ includes K 2D offset vectors from the person center to each keypoint, that is $\displaystyle T = [o^{0}_{p,x},o^{0}_{p,y},\dots, o^{K-1}_{p,x},o^{K-1}_{p,y}]$. The ground truth pose parameters are denoted using $\hat{T}$. Once the network classifies $\displaystyle p$ as centre of an instance, the pose parameters are readily available to generate the complete prediction.

We adapt Eq.~\ref{eq:mdn_prob_den}  and Eq.~\ref{eq:gaussain_cpdf} such that the mixture model predicts pose parameters $T$. That is, if we have an $M$ component mixture model ($MDN_{M}$), $\displaystyle \boldsymbol{\mu}_{m}$ would represent pose parameters predicted by component $m \in M$. Then the density of the ground truth pose parameters $\displaystyle \hat{T}$ conditioned on image $\displaystyle I$ is given by Eq.~\ref{eq:offset_conditional}, where the density function $\displaystyle \phi_m$ for each mixture component is given by Eq.~\ref{eq:comp_density}. In Eq.~\ref{eq:comp_density}, $\displaystyle T_{m}(I)$ is the input dependent pose parameters generated by component $\displaystyle m$. $\displaystyle \sigma_m(I)$ is the standard deviation of the component $\displaystyle m$ in two dimensions, that is, X and Y for horizontal and vertical axes. To account for keypoints' scale difference, in our implementation of Eq.~\ref{eq:comp_density} for human pose estimation, for each keypoint we divide $\displaystyle \sigma_m(I)$ by its scale factor provided in the COCO dataset. 

% The target dimensionality $c$ in Eq. \ref{eq:comp_density} is 2 for object detection and 34 for 
% In this framework, the dimensions of $T$ are independent within each component, but the full model does not assume such independence. 

% as shown in Eq. \ref{eq:l1_offset} . In \ref{eq:l1_offset}, $\displaystyle G = [\vec{g}^{0}_{p,x},\vec{g}^{0}_{p,y}, \dots ,\vec{g}^{K}_{p,x},\vec{g}^{K}_{p,y}]$ indicates the ground truth offsets.
% \begin{equation} \label{eq:l1_offset}
%      L = \sum_{i=1}^{i=k} |\vec{g}^{i}_{p,x}  - \vec{o}^{i}_{p,x}| + |\vec{g}^{i}_{p,y}  - \vec{o}^{i}_{p,y}|
% \end{equation}

\vspace{-10pt}

\begin{equation} \label{eq:offset_conditional}
p(\hat{T}|I) = \sum_{m=1}^{M} \alpha_m(I)\phi_m(\hat{T}|I)
\end{equation}
\vspace{-5pt}
\begin{equation} \label{eq:comp_density}
\phi_m(\hat{T}|I) = \frac{1}{(2\pi)^{c/2}\sigma_m(I)^c}  exp\Big\{ - \frac{||  \hat{T} - T_{m}(I)||^{2}}   {2\sigma_m(I)^2}\Big\} 
\end{equation}

Given the conditional probability density of the ground truth in Eq. \ref{eq:offset_conditional}, we can define the regression objective as the negative log-likelihood and minimize it using stochastic gradient descent. In Eq. \ref{eq:mdn_loss}, we provide the negative log-likelihood for the pose targets generated by MDN, where $\displaystyle N$ is the number of samples in the dataset. Essentially, this loss term replaces the popular  $\displaystyle L_1$ loss for pose regression targets. Note that we implement MDN in a dense fashion. That is, density estimation is done independently at each spatial location $\displaystyle p' \in P'$. A schematic overview of the model is shown in Fig. \ref{fig:schematic}.

\vspace{-10pt}

\begin{equation} \label{eq:mdn_loss}
    L_{T} = \sum_{i=1}^{N} -\ln{\sum_{m=1}^{M} \alpha_m(I_i)\phi_m(\hat{T}_i|I_i) }
\end{equation}

We do not modify the other loss terms used in CenterNet. This includes a binary classification loss $L_C$ for each class, offset regression loss term $L_{C_{off}}$ to compensate for lost spatial precision due to downsampling, and the term $L_{T}$ loss for pose parameters. The total loss is given in Eq. \ref{eq:full_loss}:

\vspace{-5pt}

\begin{equation} \label{eq:full_loss}
    % L_{total} = L_C + 0.1L_{MDN} + L_{HM} + L_{C_{off}} + L_{KP_{off}} + 0.1 L_{wh}
    L_{total} = \lambda_{C}L_C + \lambda_{off}L_{C_{off}} + \lambda_{T} L_{T}
\end{equation}

In the case of human pose estimation, CenterNet also adds a traditional heatmap based keypoint detection and small offset regression heads to the network. This is used for further refinement at inference. These loss terms will be denoted by $L_{HM}$ and $L_{KP_{off}}$, respectively.

% Equation. \ref{eq:full_loss_refined} shows the total loss for human pose estimation with refinement heads. 

% % \begin{equation} \label{eq:full_loss_refined}
% \begin{dmath}\label{eq:full_loss_refined}
%     % L_{total} = L_C + 0.1L_{MDN} + L_{HM} + L_{C_{off}} + L_{KP_{off}} + 0.1 L_{wh}
%     L_{total_refined} = \lambda_{C}L_C + \lambda_{off}L_{C_{off}} + \lambda_{T} L_{T} +
%     \lambda_{off}L_{KP_{off}} + \lambda_{HM}L_{HM} 
% \end{dmath}    
% % \end{equation}

% In the total loss  formula, $\lambda_{x}$ is the weight for loss term $x$.

In our experiments, we have $\lambda_{T}=0.1$. It is tuned such that the performance of $MDN_1$ is on par with that of CenterNet (which is a single-mode model). Other loss weights are the same as the ones used in CenterNet, that is $\lambda_{C}=1$, $\lambda_{off}=0.1$, and $\lambda_{HM} = 1$.

%Assuming that $\displaystyle \{\textbf{p}_0 \dots \textbf{x}_n\}$ is the set of all location in the image and that 

\subsection{Inference}
Once the network is trained, at each spatial location, the classification branch determines if it is the center of an instance (we use the bounding box center for ground truth). If yes, we can use either the mixture of outputs by the components or the one with the highest score to generate the pose parameters. We tried both cases and found out that using the maximum component leads to slightly better results.

% CenterNet 1X and 2x are trained by us under the same condition used for training $MDN_3$
\vspace{-5pt}
\section{Experiments}

We have conducted extensive experiments on the MS COCO 2017 dataset \cite{lin2014microsoft}. From the training split (\textit{coco-train}), we use all the 118k images for object detection and the images with at least one person instance (64k images) for human pose estimation. To compare against the state-of-the-art, we use the COCO test-dev split (\textit{coco-test-dev}). All other evaluations are done on COCO validation split which contains 5k images (\textit{coco-val}). The evaluation metric in all experiments is COCO average precision (AP). For object detection we experiment using hourglass-104 (HG) \cite{law2018cornernet} and deep layer aggregation (DLA34) \cite{yu2018deep} architectures. For human-pose estimation, we only use HG.
%  and submit the results to the official evaluation server.

% The backbone network is based on a version of the stacked hourglass network \cite{newell2016stacked} presented in \cite{law2018cornernet}. DLA-34  This backbone is also used by CornerNet \cite{law2018cornernet}, ExtremeNet \cite{zhou2019bottom}, and our baseline model CenterNet.

% It has 104 convolution layers, roughly organized in two initial convolutional layers, two hourglass models (each with five levels), and two convolutional layers before prediction heads.
% We refer to this architecture as LargeHG.
%, we have used a smaller version of this model called SmallHG architecture obtained by replacing the residual layers with conv layers, and removing one layer from each hourglass level. Unless specified otherwise, all models are trained based on LargeHG architecture.  

\subsection{Training}
We use the ADAM optimizer \cite{kingma2014adam} for training our models. For models based on HG, we use batch size 12 and three different schedules taking 50 (1X), 100(2X), and 150(3X) epochs. We initialize the learning rate to 2.5e-10 and drop it by factor 10 at the tenth epoch from the last. For the ones based on DLA34, we train for 140 epochs (1X) with batch size 32 and learning rate 2e-4, which is dropped by factor 10 at epochs 90 and 120. These settings are analogous to the ones used in CenterNet \cite{zhou2019objects}. Unless indicated otherwise, for all experiments, we use HG with the 1X schedule. To allow for proper comparisons, we train all models, including the single-mode base model, CenterNet, from scratch.

% In training for pose estimation, for computing the likelihood, we ignore the keypoints that are not annotated. 

\subsection{Activation function for variance terms}
% Proper initialization of the Gaussian variance terms is very important for successful training of the network.
A typical formulation for Gaussian MDNs uses an activation function for variance terms that generates positive values \cite{cui2019multimodal,rupprecht2017learning,li2019generating}. However, this causes numerical instabilities when some of the mixture components are redundant or do not have a significant contribution. Unlike the significant modes, the variance term of these components do not get trained to be in a suitable range, and may lie between zero and one. If only a small portion of samples get assigned to them at training, highly irregular gradients will be generated, impeding proper training of the model. For example, average precision on \textit{coco-val} can fluctuate between 5 and 30 percent in consecutive epochs even after being trained for tens of epochs. A simple remedy is to prevent the variance term from falling in the range $(0,1)$. Hence, we use a modified version of exponential linear units (ELU) \cite{clevert2015fast} for activation of the variance terms such that the minimum value is one. We experimented with larger minimum values (up to 10) but did not observe any significant difference. 

% in Table \ref{fig:initi_stability} we show accuracy of a mixture model with five components in the middle of training with minimum variance value of 0. The table also includes the distribution of components at each epoch to show how unused component lead to this instability.

% To avoid other possible numerical issues, we have implemented the log likelihood using the LogSumExp function. 

\subsection{Object Detection}

In table \ref{fig:valres_object} we provide evaluations on the \textit{coco-val} set for the baseline and mixture models with two to five components on HG architecture. With the 1X schedule, $MDN_3$ achieves an impressive 3.1 percentage points improvement. 

\begin{figure}[t]
\centering
\resizebox{.5\textwidth}{!}{
%\framebox[4.0in]{$\;$}
\includegraphics[width=1\textwidth]{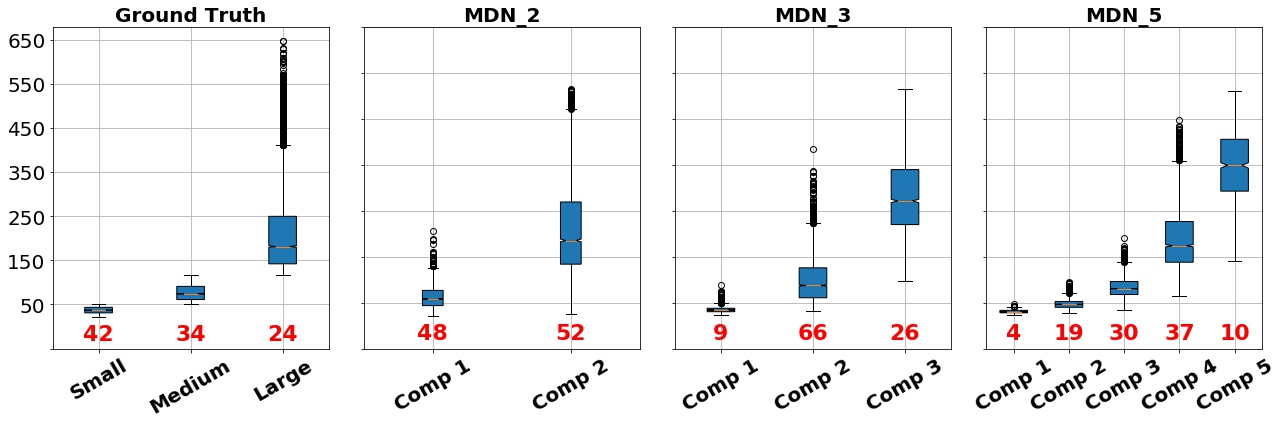}
}
\captionof{figure}{Distribution of ground truth (GT) object scales compared to the distribution of scale by mixture models trained on HG architecture. Components are named in order of their scale range. Below the boxes, we also indicate the distribution of GT instances and mixture components.}
\label{fig:scale_distribution}
\end{figure}

\vspace{-10pt}
\subsubsection{Analysis of the components}

To gain insight into what each component learns, we visually investigated the objects regressed by each component (samples can be seen in Fig. \ref{fig:object_sample}). It turns out that MDN separates the dataset based on the objects' scale. To confirm this in quantitative terms, we looked at the distribution of scale for objects regressed by different components. When we compare the scale distribution of the ground truth data, we observe a strong correlation, as shown in Fig.~\ref{fig:scale_distribution}. Quantitatively, the Pearson correlation coefficients for maximum component and the scale on \textit{coco-val} is $0.76$. The correlation between components and categories is only $0.04$. In other words, there is no obvious relationship between components and categories.

As a further test, we trained a version of the base model with separate box prediction heads for each category. However, we did not observe any notable difference compared to the case where there is only a shared box prediction head. Therefore, merely using category-specific prediction heads does not have any benefit.

Based on table \ref{fig:valres_object}, the number of components does not seem to have a significant effect when increased to more than three. However, according to Fig. \ref{fig:scale_distribution}, it does lead to better separation of the data. The fact that $MDN_5$ yields better separation, but not higher accuracy, seems odd. We believe this could be because the classification branch is not keeping up with the regression head. This is an interesting question for future research; is it possible to achieve even higher accuracy with an increased number of components?

\begin{table}
\centering
\resizebox{.45\textwidth}{!}{
\begin{tabular}{c|c|c|c|c|c|c|c}
\textbf{Model} & \textbf{Comp.} & \textbf{AP} & \textbf{AP_{50}} & \textbf{AP_{75} }& \textbf{AP_{S} }& \textbf{AP_{M}} & \textbf{AP_{L}} \\ \hline

Single mode base & - & 35.9 & 54.5 & 38.4 & 20.4 & 39.5 & 46.1 \\ \hline 
\multirow{3}{*}{\pmb{$MDN_{2}$}} & All & 38.4 & 56.6 & 41.2 & 22.3 & 42.4 & 49.9 \\ \cline{2-8}
& 1 & 27.8 & 42.6 & 28.6 &  8.0 & 30.2 & 45.8  \\ 
& 2 & 28.0 & 52.3 & 26.9 & 22.2 & 38.5 & 27.8 \\ 
\hline
%  \cline{2-12}
\multirow{4}{*}{\pmb{$MDN_{3}$}} & All & \textbf{39.0} & \textbf{57.1} & \textbf{42.1 }& 22.1 &\textbf{ 43.0} & \textbf{50.1}\\ \cline{2-8}
& 1 & 12.7 & 31.9 &  8.2 & 15.7 & 20.0 &  6.6  \\ 
& 2 & 35.4 & 53.4 & 38.4 & 19.5 & 42.0 & 45.8  \\ 
& 3 & 22.2 & 34.7 & 22.6 &  2.9 & 20.9 & 39.8\\ 
\hline
\multirow{6}{*}{\pmb{$MDN_{5}$}} & All & 38.8 & 56.8 & 41.7 & \textbf{22.2} & 43.0 & 49.8\\ \cline{2-8}
& 1 & 24.0 & 42.9 & 23.7 & 16.7 & 40.1 & 24.4 \\
& 2 & 1.8 &  5.2 &  1.0 &  4.3 &  0.4 &  0.0 \\
& 3 & 10.6 & 22.1 &  9.3 & 20.9 & 13.9 &  1.0\\
& 4 & 27.3 & 40.8 & 28.6 &  6.4 & 30.8 & 45.3 \\
& 5 & 11.1 & 18.4 & 10.8 &  0.1 &  6.2 & 22.3 \\ 
\end{tabular}
}
\caption{Object detection evaluation. For mixture models, we evaluate with the full model and separately for each component (Comp.) when it is used to make all predictions. All models are trained on HG for 1X from scratch.}
\label{fig:valres_object}
\end{table}

\begin{figure}[t]
\centering
\subfloat[Various architectures]{\label{fig:speedaccfull}\includegraphics[width=.3035\textwidth]{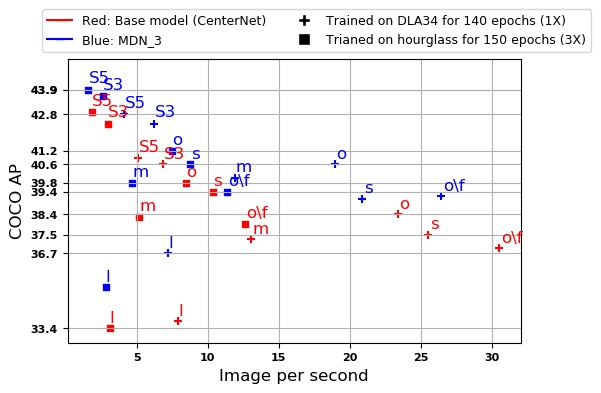}}
\subfloat[Various training input sizes]{\label{fig:speedacc10}\includegraphics[width=.2\textwidth]{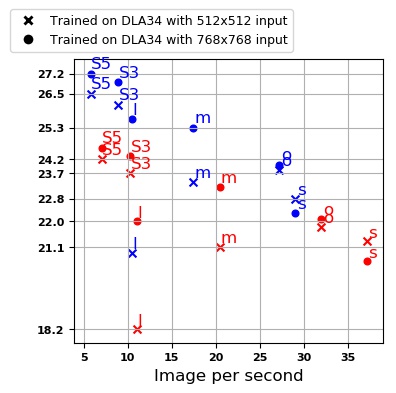}}

\caption{Speed accuracy trade-off. \textbf{s}, \textbf{m}, \textbf{l}, and \textbf{o} indicate test time input size of 512, 768, 1024, and original input size, respectively. \textbf{S5} and \textbf{S3} indicate multi scale test with 5 and 3 scales, respectively. Due to limited resources for training on larger input size, models in \protect\subref{fig:speedacc10} are trained for 100 epochs on 10\% of coco-train (sampled uniformly at random.). Evaluations in \protect\subref{fig:speedaccfull} use left-right flip, except if marked with \textbf{\textbackslash f}. Evaluations in \protect\subref{fig:speedacc10} are done without left-right flip. Best seen in color and zoomed-in.}
\label{fig:speedacc}
\end{figure}

\begin{figure}[t]
\centering
\resizebox{.35\textwidth}{!}{
%\framebox[4.0in]{$\;$}
\includegraphics[width=1\textwidth]{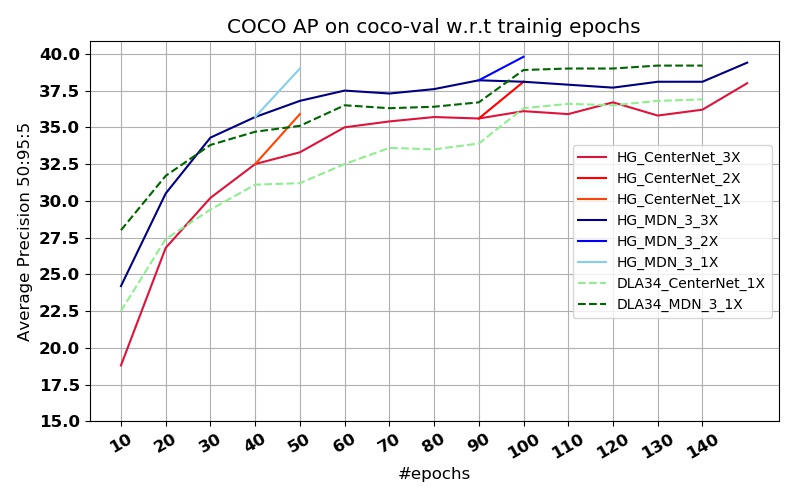}
}
\caption{Convergence of $MDN_{3}$ vs. single-mode base.
%(CenterNet). 
% CenterNet4X reaches AP=0.403 with a 4X schedule. $MDN_{3}$ reaches AP=0.400 with 2X schedule.
}
\label{fig:convergence}
\end{figure}

\vspace{-5pt}
\subsubsection{Speed-accuracy trade-off}
In Fig. \ref{fig:speedacc} we show the speed-accuracy trade-off for $MDN_3$ compared to the base single mode model. According to Fig. \ref{fig:speedacc} \subref{fig:speedaccfull}, $MDN_3$ consistently achieves higher accuracy at higher speed. In particular, with 5 scale evaluation, MDN on DLA34 is as accurate as the CenterNet on stronger HG architecture, however twice faster (see red square labeled \textit{S5} vs. blue cross labeled \textit{S5}). 

Given that MDN divides data based on object scale, we hypothesize that training on larger input should result in even better accuracy. So, on  DLA34, we train CenterNet and $MDN_{3}$ on input size 768x768 (up from default 512x512). However, since training on larger input demands almost twice GPU memory, we train for 100 epochs on a 10\% subset of \textit{coco-train} sampled uniformly at random. This is solely due to the limited computational resources we have access to. Fig. \ref{fig:speedacc} shows that \subref{fig:speedacc10}, when trained on larger input, $MDN_{3}$ evaluated at single scale at resolution 768x768 surpasses accuracy of the base model evaluated with 5 scales, while being more than twice faster with FPS 17,47 vs. 7.04 (see red circle labeled \textit{S5} vs. blue circle labeled \textit{m}).

\vspace{-15pt}
\subsubsection{Convergence}

% The official CenterNet model, publicly published by the authors~\footnote{https://github.com/xingyizhou/CenterNet}, is fine-tuned on top of ExtremeNet \cite{zhou2019bottom} for 50 epochs. As we investigated, ExtremeNet itself is fine-tuned on top of CornerNet \cite{law2018cornernet} for 50 epochs. And, CornerNet is trained from scratch for 100 epochs. It is safe to say that official CenterNet is effectively trained for 200 epochs. It is a very long schedule, but for object detection, it is common practice to train for such long schedules \cite{redmon2016you}. We call this schedule 4X and refer officially published model as CenterNet4X. In our experiments on two Nvidia Tesla P100 GPUs, an epoch of training takes more than three hours. Therefore training for long schedules is extremely costly in terms of time and energy consumption.

In Fig. \ref{fig:convergence} we illustrated that an important aspect of using a mixture model is a faster convergence rate. For both HG and DLA34 architectures, $MDN_3$ gains higher accuracy more quickly than the single-mode base. The base model gradually recovers some of the gap, but never reaches the same accuracy level.  

% While $MDN_3$ reaches an AP of $39.0$ with 1X schedule, CenterNet4X with four times longer training is only $1.3$ better in terms of accuracy, with an AP of $40.3$. With 2X schedule, $MDN_3$ reaches an AP of $40.0$ while CenterNet achieves $38.1$ and CenterNet4X gets to $40.03$. That is only $0.03$ better than $MDN_3$, which can partially be attributed to randomness in the training procedure, according to the authors. We trained $MDN_3$ for a longer schedule, but it does not provide any significant improvement.  Please note that except for CenterNet4x, we have trained all other versions of CenterNet from scratch under the same conditions as we have used for training our mixture models. 

\vspace{-10pt}
\subsubsection{Comparison to the state-of-the-art}
\vspace{-10pt}
The essence of our contribution is to improve object detection by incorporating it into a mixture model. It is crucial to conduct experiments in a way that the real effect of the new formulation is easily understood. However, the official CenterNet model (the single-mode baseline) is trained for 50 epochs after initializing from ExtremeNet \cite{zhou2019bottom}. As we investigated, ExtremeNet itself is fine-tuned on top of CornerNet \cite{law2018cornernet} for 50 epochs. And, CornerNet is trained from scratch for 100 epochs. It is safe to say that the model is trained for about 200 epochs (CenterNet4x). This schedule makes it difficult, if not impossible, to make a proper comparison and measure the effectiveness of the mixture formulation.  Therefore, for comparisons on \textit{coco-test-dev} too, we train both our mixture model and the base from scratch.

%  is possible only if we train base model under the same setting. Hence, for comparison to the state of the art

% To minimize any potential error, we use the publicly available official CenterNet source code.  

In table \ref{fig:state_of_the_art_object} we provide evaluation results along with run-time speed for our model and the single-stage state-of-the-art models on \textit{coco-test-dev}. $MDN_3$ significantly improves the baseline model. Its precision is slightly lower than the recent FSAF model \cite{zhu2019feature}, however it is much faster. Please note that CenterNet4x achieves test AP of $42.1$, slightly better than the AP of $41.5$ achieved by $MDN_3$.

\begin{table}[h]
\centering
\resizebox{.48\textwidth}{!}{
\begin{tabular}{c|c|c|c|c|c|c|c}
% \multicolumn{1}{c}{\thead{model}}&\multicolumn{1}{c}{FPS}&\multicolumn{1}{c}{AP} &\multicolumn{1}{c}{AP_{50}} &\multicolumn{1}{c}{AP_{75}} &\multicolumn{1}{c}{AP_{S}} &\multicolumn{1}{c}{AP_{M}} &\multicolumn{1}{c}{AP_{L}} \\ \hline

\textbf{Model} & \textbf{FPS} & \textbf{AP} & \textbf{AP_{50}} & \textbf{AP_{75} }& \textbf{AP_{S} }& \textbf{AP_{M}} & \textbf{AP_{L}} \\ \hline
% MaskRCNN \cite{he2017mask} & 39.8 & 62.3 & 43.4 & 22.1 & 43.2 & 51.2 \\
% SNIPER \cite{singh2018sniper} & 46.1 & 67.0 & 51.6 & 29.6 & 48.9 & 58.1 \\
% PANet \cite{liu2018path} & 47.4 & 67.2 & 51.8 & 30.1 & 51.7 & 60.0 \\
% TridentNet \cite{li2019scale} & 48.4 & 69.7 & 53.5 & 31.8 & 51.3 & 60.3
% \hline
YOLOv3 \cite{redmon2018yolov3}& 20 & 33.0 & 57.9 & 34.4 & 18.3 & 25.4 & 41.9 \\
Gaussian YOLOv3 \cite{choi2019gaussian}& - & 36.1 & - & - & - & - & - \\ 
RetinaNet \cite{lin2017focal}& 5.4 & 40.8 & 61.1 & 44.1 & 24.1 & 44.2 & 51.2 \\
CornerNet \cite{law2018cornernet}& 4.1 & 40.5 & 56.5 & 43.1 & 19.4 & 42.7 & 53.9 \\
ExtremeNet \cite{zhou2019bottom}& 3.1  & 40.2 & 55.5 & 43.2 &  20.4 & 43.2 & 53.1 \\
FSAF \cite{zhu2019feature}& 2.7 & 42.9 & 63.8 & 46.3 & 26.6 & 46.2 & 52.7 \\
% \hdashline
% CenterNet (4X schedule) \cite{zhou2019objects} & 8.44 & 42.1 & 61.1 & 45.9 & 24.1 & 45.5 & 52.8 \\
% \hdashline
% CenterNet (2X schedule on HG)& 8.44 & 40.0 & 58.7 & 43.5 & 22.9 & 43.3 & 49.3 \\
% \pmb{$MDN_{3}$} (ours) (2X schedule on HG)&  7.47 & 41.5 & 59.9 & 45.1 & 24.1 & 45.0 & 51.5\\
\hdashline
Single mode base on HG& 8.44 & 0.401 & 0.589 & 0.436 & 0.232 & 0.435 & 0.496 \\
\pmb{$MDN_{3}$} (ours) on HG&  7.47 & 0.416 & 0.599 & 0.452 & 0.239 & 0.451 & 0.518 \\
\hdashline
Single mode base on DLA34& 23.29 & 0.386 & 0.566 & 0.418 & 0.190 & 0.425 & 0.505 \\
\pmb{$MDN_{3}$} (ours) on DLA34&  18.97 & 0.406 & 0.582 & 0.438 & 0.206 & 0.441 & 0.538 \\

\end{tabular}
}
\caption{Comparison to bottom-up state-of-the-art object detectors on \textit{coco-test-dev}. Results correspond to single-scale inference with left-right flip augmentation at test time. Models on HG are trained for 3X and on DLA34 for 1X. }
\label{fig:state_of_the_art_object}
\end{table}

\begin{table}
\centering
\resizebox{.4\textwidth}{!}{
\begin{tabular}{c|c|c|c|c|c|c}
% \hline
% \multicolumn{1}{c}{\thead{Model}}&\multicolumn{1}{|c}{\thead{Component}}&\multicolumn{1}{|c}{AP} &\multicolumn{1}{c}{AP_{50}} &\multicolumn{1}{c}{AP_{75}} &\multicolumn{1}{c}{AP_{M}} &\multicolumn{1}{c}{AP_{L}} \\ \hline

\textbf{Model} & \textbf{Comp.} & \textbf{AP} & \textbf{AP_{50}} & \textbf{AP_{75} }& \textbf{AP_{M}} & \textbf{AP_{L}} \\ \hline

% \thead{Single Mode\\ using L2 loss} 
% \cite{zhou2019objects}
Single mode base & - & 46.4 & 75.8 & 50.3 & 43.9 & 53.2 \\ \hline 

\multirow{3}{*}{\pmb{$MDN_{2}$}} & All & \textbf{52.3} & \textbf{78.2} & 57.2 & 50.0 & 58.9 \\ \cline{2-7}
& 1 & 37.3 & 72.0 & 34.6 & 38.0 & 40.0 \\ 
& 2 & 37.7 & 62.6 & 39.1 & 33.4 & 46.2\\ 
\hline
%  \cline{2-12}
\multirow{4}{*}{\pmb{$MDN_{3}$}} & All & \textbf{52.3} & 77.2 & \textbf{58.0} & \textbf{50.6} & \textbf{59.3 }\\ \cline{2-7}
& 1 & 34.6 & 54.5 & 37.0 & 30.5 & 42.8 \\ 
& 2 & 40.7 & 70.9 & 42.0 & 39.0 & 46.8 \\ 
& 3 & 23.9 & 60.6 & 14.9 & 27.3 & 24.0 \\ 
\end{tabular}
}
\caption{Human pose estimation evaluation. For mixture models, we show the results for the full model and separately for each component when it is used to make all predictions. Models are trained on HG for 1X from scratch.}
\label{fig:valres_pose}
\end{table}

\vspace{-5pt}
\subsection{Human pose estimation}
\begin{figure*}[h]
\centering
%\framebox[4.0in]{$\;$}
\subfloat[Sample object detection by \pmb{$MDN_{3}$}. The three different modes are color-coded (blue, green, and red)]{\label{fig:object_sample}\includegraphics[width=.81\textwidth]{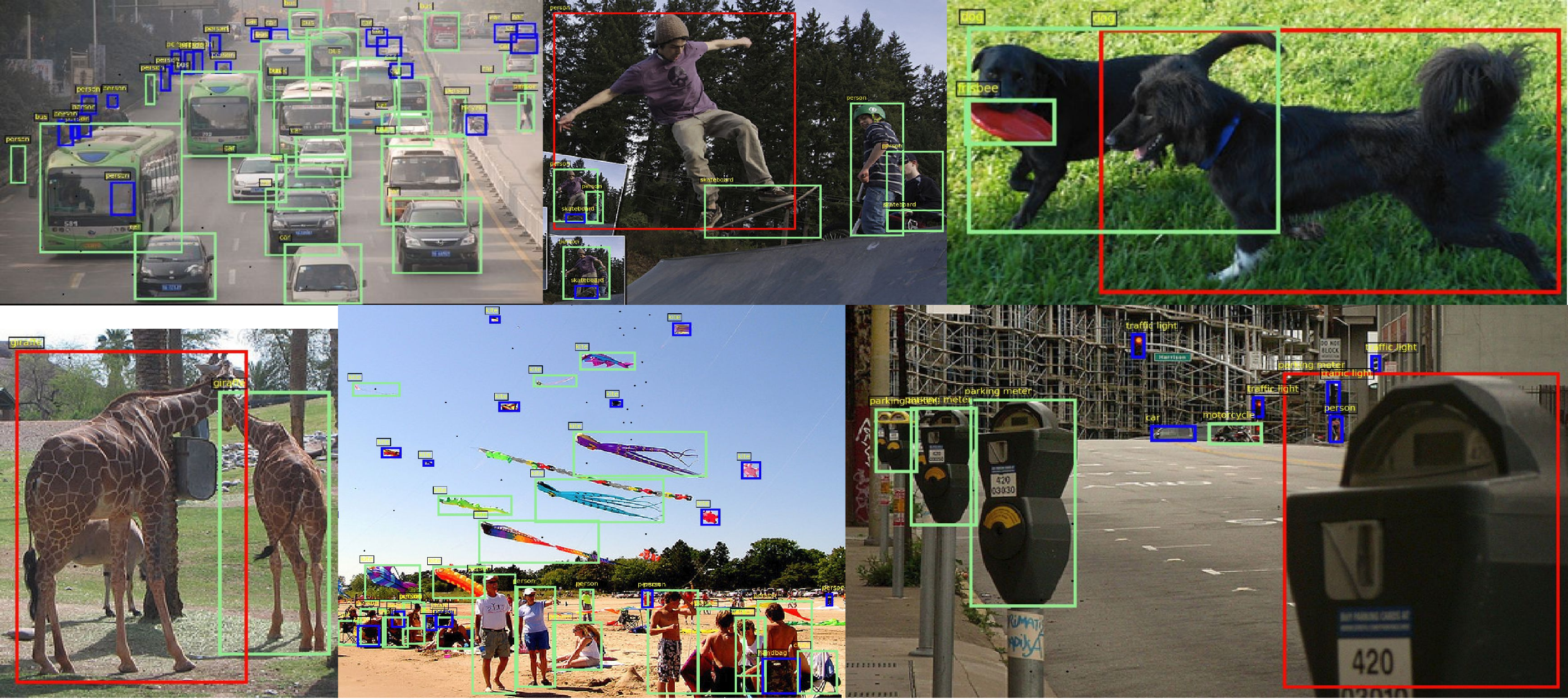}}

\subfloat[Sample pose estimation by \pmb{$MDN_{3}$}. Ellipses represent uncertainty ($\sigma_m$ in Eq. \ref{eq:comp_density}). The modes are color-coded using the bounding box color.]{\label{fig:pose_sample}\includegraphics[width=.81\textwidth]{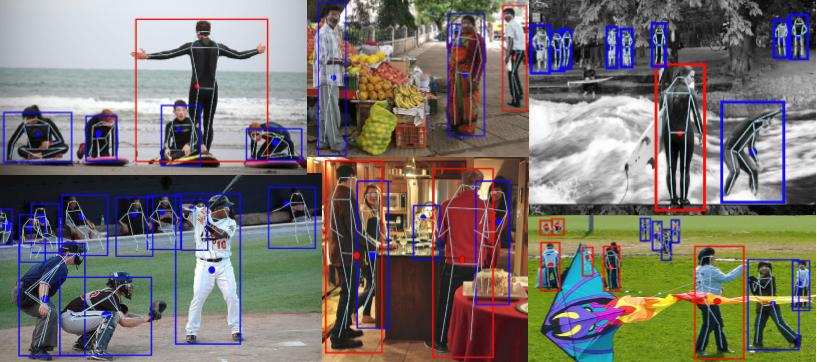}}

\caption{Sample predictions on \textit{coco-val}. We provide more visualizations in the supplementary material.}
\label{fig:vis_sample}
\end{figure*}

Table \ref{fig:valres_pose} shows the evaluation results for mixture models trained for human pose estimation on COCO. The mixture model leads by a significant margin in terms of AP. However, the interpretation of components is different than that of object detection. Here, only two significant modes are retrieved, no matter how many components we train. We experimented with up to 10 components and observed that having more than two components results in slightly better recall ( $< 1\%$), but does not improve precision. By visualizing the predictions, it becomes clear that one of the modes focuses on roughly frontal view instances and the other one on instances with a backward view. Fig. \ref{fig:pose_sample} shows sample visualisation from $MDN_3$ on HG trained with 3X schedule. In the following, we provide more analysis regarding how the input space gets divided by the mixture model.
\vspace{-10pt}
\subsubsection{Analysis of the components}
Gaussian mixture models enable us to analyze the model based on its uncertainty about samples and compute informative statistics of the modes. To this end, we analyze the difference among components of $MDN_3$ in terms of uncertainty. For the predictions made by each component, we compute the mean and standard deviation of the Gaussian variance terms, which quantifies uncertainty. The statistics, presented in table \ref{fig:stats_variance_pose}, show that indeed, there is a significant difference in the mean of the variance term for different components. We believe this difference is the actual reason behind dividing the data based on viewpoint. We also see that the second mode accounts for cases with more considerable uncertainty. But, it is responsible for only a negligible number of predictions. 

\begin{table}
\centering
\resizebox{.49\textwidth}{!}{
\begin{tabular}{c|c|c|c||c|c|c|c}
% \hline
    % \multicolumn{1}{c}{\thead{Component}} & \multicolumn{1}{c}{\thead{Prediction rate}} & \multicolumn{1}{c}{\thead{ mean (STD) \\variance in X axis}} & \multicolumn{1}{c}{\thead{mean (STD)\\ variance in Y axis}} 
% &&& &\multicolumn{2}{c}{\thead{Left body parts}} & \multicolumn{2}{c}{\thead{Right body parts}} \\

% \multicolumn{1}{c}{\thead{Component}} & \multicolumn{1}{c}{\thead{Prediction rate}} &
% \multicolumn{1}{c}{\thead{ mean (STD) \\variance in X axis}} & \multicolumn{1}{c}{\thead{mean (STD)\\ variance in Y axis}} & \multicolumn{1}{c}{\thead{X axis\\ (left parts)}} & \multicolumn{1}{c}{\thead{Y axis \\ (left parts)}} & \multicolumn{1}{c}{\thead{X axis \\ (right parts)}} & \multicolumn{1}{c}{\thead{Y axis \\ (right parts)}}\\
\thead{\textbf{Component}} & \thead{\textbf{Prediction rate}} &
\thead{ \textbf{mean (std)} \pmb{$\sigma_{x}$}} & \thead{\textbf{mean (std)} \pmb{$\sigma_{y}$}} & \thead{\pmb{$O^{+}_{x}$} \textbf{rate}\\ \textbf{(Left)}} & \thead{\pmb{$O^{+}_{y}$} \textbf{rate} \\  \textbf{(Left)}} & \thead{\pmb{$O^{+}_{x}$} \textbf{rate} \\  \textbf{(Right)}} & \thead{\pmb{$O^{+}_{y}$} \textbf{rate} \\  \textbf{(Right)}}\\ \hline
1 (Front) & 73.1\% & 58 (52) &   61 (57) & 85\% & 51\% & 25\% & 50\% \\
2 ( - ) & 0.3\% & 803 (1106) &   547 (846) & 57\% & 51\% & 49\% & 55\%  \\
3 (Back)& 26.6\% & 76 (119) & 79 (120) & 27\% & 48\% & 83\% & 47\%  \\
\end{tabular}
}
\caption{For predictions made by each component of $MDN_{3}$ table shows the mean and standard deviation (std) of the variance terms in X and Y axes ( $\sigma_{x}$ and $\sigma_{y}$), and ratio of mean vectors (offset vectors) in the positive direction of X and Y axes ($O^{+}_{x}$  and $O^{+}_{y}$ ) for left and right parts.}
\label{fig:stats_variance_pose}
\end{table}

Now let us look into some statistics of the mean vectors (i.e., offset vector) predicted by each mixture component. We divide the body parts into two sets of all the left and all the right parts. Then for each set, compute the portion of vectors in the positive direction of horizontal and vertical axes. We expect to see a significant difference in the direction of vectors along the horizontal axis between components and also between the left and right parts. Remember that vectors point from the body center to each keypoint, so if we go from the front view to the back view, the direction will be flipped in the horizontal axis. In table \ref{fig:stats_variance_pose}, we see a considerable difference in the horizontal direction of vectors for left and right parts between front and back components. This seems to be the reason for viewpoint turning out to be the dominant factor recovered by the mixture model.

We further compare the distribution of the samples in the dataset w.r.t. face visibility and occlusion of keypoints against the distribution of predictions by components of $MDN_2$. We use the nose visibility as a rough indicator of face visibility. As shown in table \ref{fig:comp_stats}, the prediction distribution correlates well with the face visibility, which is an indicator of viewpoint in 2D. The majority of instances in the dataset are in frontal view, and similarly, the front view component makes the majority of the predictions. Related to our results, \cite{belagiannis2017recurrent} have shown that excluding occluded keypoints from training leads to improved performance. More recently, \cite{ye2018occlusion} achieves more accurate 3D hand pose estimations by proposing a hierarchical model for estimation occlusion and using it to choose the right downstream model for making the final prediction. 
% predicts occlusion of a keypoint to use it for selecting a downstream model. 
% And, here we illustrate that occlusion caused by viewpoint imposes more challenge to spatial regression models, than other possible factors, like variation in pose itself.

% COCO validation split by ignoring predictions by each of components or forcing all predictions to be made by a particular component. The detailed evaluations are presented in table \ref{fig:occlision_analysis}. The results show that the components correlate well with face viability, confirming the conclusion we make by visualising predictions. 

\begin{table}[t]
\centering
\resizebox{.48\textwidth}{!}{
\begin{tabular}{c|c|c|c|c|c}
% \multicolumn{1}{c}{\thead{Occluded\\ Keypoints} } &\multicolumn{1}{c}{\thead{Visible \\ Keypoints}} &\multicolumn{1}{|c}{ \thead{Occluded Face}} &\multicolumn{1}{c}{\thead{Visible  Face}}   &\multicolumn{1}{|c}{ \thead{ Back mode} }  &\multicolumn{1}{c}{ \thead{Front mode}} \\ \hline 

\thead{\textbf{Occluded}\\ \textbf{Keypoints}} & \thead{\textbf{Visible} \\ \textbf{Keypoints}} & \thead{\textbf{Occluded Face}} & \thead{\textbf{Visible  Face}} & \thead{\textbf{Back mode}}  &  \thead{\textbf{Front mode}} \\ \hline 

  12.3 & 87.7 & 30.3 (22.1) & 69.7 (77.9)  & 25.5 (27.0) & 74.5 (70.0)
\end{tabular}
}
\caption{Distribution of ground truth face and keypoints visibility compared to that of maximum mixture component. Statistics are based on GT instances with more than 5 annotated keypoints (or $\geq10$), and model predictions with the score at least 0.5 (or $\geq0.7$). "visible face" includes instances with the visible and annotated nose, and "occluded face" those with the occluded or un-annotated nose.}
\label{fig:comp_stats}
\end{table}

\vspace{-10pt}
\subsubsection{Fine-grained evaluation}
To understand what body parts gain the most from the MDN, we do a fine-grained evaluation for different subsets of keypoints. We modify the COCO evaluation script such that it only considers keypoints we are interested in. Table \ref{fig:per_kp} shows the results. The first three rows of the table illustrate the sensitivity of the COCO evaluation metric. For the facial keypoints, where the metric is the most sensitive, the improvement is larger. However, the biggest improvement comes for the wrists, which have the highest freedom to move. On the other hand, for torso keypoints, which are the most rigid, there is almost no improvement.

 \begin{table}
    \centering
    \resizebox{.49\textwidth}{!}{
        \begin{tabular}{c|ccc|ccccccccc}
             & \textbf{All} & \textbf{Facial} & \textbf{Non-Facial}  & \textbf{Nose}  & \textbf{Ears} & \textbf{Shoulders}& \textbf{Wrists} & \textbf{Hips} & \textbf{Ankles}  \\ \hline
             
             GT Disp. 1 & 96.0 & 83.7 & 99.6 & 77.9 & 88.7 & 99.4  & 97.1 & 99.2& 95.2  \\
             GT Disp. 2 & 80.4 & 47.4 & 93.4 & 42.3 & 59.8 & 93.4 & 86.8 & 97.2& 91.7  \\ 
             GT Disp. 3 & 63.0 & 25.6 & 82.7 & 22.6  & 35.8 & 82.0  & 71.2& 91.3 & 83.3  \\
            \hline                 
            Base model & 46.4 & 44.3 & 45.7 & 42.3 & 43.9 & 59.5  & 31.7 & 58.5 & 38.2\\
            \pmb{$MDN_{2}$} & 52.3 & 54.1 & 49.9 & 50.3 & 54.0 & 60.3 & 41.7 & 58.3 & 42.6 \\ 
            % \thead{ Relative\\improvement} & 12.7 \%&  22.1\% &  9.2 \% & 18.9\%  &  23.0\% &  1.3\% &  31.5\% &  -0.3\%&  11.5\% 
        \end{tabular}    
        }
    \caption{Fine-grained evaluation. \textit{GT Disp. x} means ground truth is displaced by \textit{x} pixels in random direction.}
    \label{fig:per_kp}
\end{table} 

\vspace{-10pt}
\subsubsection{Comparison to the state-of-the-art}
Table \ref{fig:state_of_the_art_pose} compares our mixture model to the state-of-the-art on \textit{coco-test-edv}. Analogous to our object detection setting, in order to make a proper comparison to CenterNet, we train it along with our mixture model from scratch with the 3X schedule on HG. The official CenterNet for human pose estimation is also trained with a 3X schedule but with different batch-size. Due to that, we observe a discrepancy in the results; we get an AP of 61.8 on the test server, but the official model gets 63. When no refinement is used, our training gets AP of 55.6 while the official model gets 55.0. Therefore, for the sake of fair comparison, in table \ref{fig:state_of_the_art_pose}, we only display the results obtained by our own training.

\begin{table}[h]
\centering
\resizebox{.48\textwidth}{!}{
\begin{tabular}{c|c|c|c|c|c|c}
% \multicolumn{1}{c}{\thead{model}}&\multicolumn{1}{c}{FPS}&\multicolumn{1}{c}{AP} &\multicolumn{1}{c}{AP_{50}} &\multicolumn{1}{c}{AP_{75}} &\multicolumn{1}{c}{AP_{M}} &\multicolumn{1}{c}{AP_{L}} \\ \hline

\textbf{Model} & \textbf{FPS} & \textbf{AP} & \textbf{AP_{50}} & \textbf{AP_{75} }& \textbf{AP_{M}} & \textbf{AP_{L}} \\ \hline

% MSPN \cite{li2019rethinking} & 76.1 & 93.4 & 83.8 & 72.3 & 81.5 \\
% Mask R-CNN \cite{he2017mask} & 63.1 & 87.3 & 68.7 & 57.8 & 71.4 \\
% Simple baselines \cite{xiao2018simple} & 73.7 & 91.9 & 81.1 & 70.3 & 80.0  \\
% CPN+ \cite{chen2018cascaded} & 73.0 & 91.7 &80.9 & 69.5 & 78.1 \\
% \hline
CMU-Pose \cite{cao2017realtime} & 12.98 & 61.8 & 84.9 & 67.5 & 58.0 & 70.4  \\
AssociativeEmbedding \cite{newell2017associative} & 9.3 & 62.8 & 84.6 & 69.2 & 57.5 & 70.6 \\
PersonLab \cite{papandreou2018personlab} & 2.15 & 66.5 & 88.0 & 72.6 & 62.4 & 72.3  \\
\hline
% Base model (CenterNet) \cite{zhou2019objects}& 55.0 & 83.5 & 59.7 & 49.4 & 64.0\\
Single-mode base & 7.65 & 55.6 & 82.8 & 61.1 & 49.6 & 65.9 \\
%CenterNet w/o left-right flip (ours)  & 54.6 & 80.7 & 60.1 & 48.7 & 63.5 & 12.89 \\
\pmb{$MDN_{3}$} (ours) &  7.26 & 57.9 & 82.7 & 63.7 & 52.3 & 67.8   \\
\hdashline
Single-mode base w/o flip & 12.79 & 56.0 & 82.6 & 61.6 & 52.6 & 63.7  \\
% Base model w/o flip \cite{zhou2019objects}  & 54.6 & 80.7 & 60.1 & 48.7 & 63.5 \\
\pmb{$MDN_{3}$} w/o flip (ours) & 11.77 & 59.0 & 82.7 & 65.3 & 56.4 & 65.9  \\
\hdashline
Single-mode base refined & 7.13 & 61.8 & 85.4 & 68.0 & 57.4 & 70.5 \\
% Base model refined \cite{zhou2019objects}  & 63.0 & 86.8 & 69.6 & 58.9 & 70.4   \\ 
\pmb{$MDN_{3}$} refined (ours) & 7.04 & 62.9 & 85.1 & 69.4 & 58.8 & 71.4

\end{tabular}
}
\caption{Comparison to the state-of-the-art bottom-up human pose estimators on \textit{coco-test-dev}. Evaluations are done at single-scale, and with left-right flip. $MDN_{3}$ and the base model are trained on HG for 3X.}
% The top rows belong to top-down models, and the middle rows  to bottom-up models.

\label{fig:state_of_the_art_pose}
\end{table}

% \begin{figure*}[h]
% \centering
% %\framebox[4.0in]{$\;$}
% \includegraphics[width=.9\textwidth]{images/object_sample.pdf}
% \caption{Sample predictions by \pmb{$MDN_{3}$} on \textit{coco-val}. The three different modes are color-coded (blue, green, and red). More visualizations are provided in the supplementary material.}
% \label{fig:object_sample}
% \end{figure*}

% \begin{figure*}[h]
% \centering
% %\framebox[4.0in]{$\;$}
% \includegraphics[width=.9\textwidth]{images/pose_sample.pdf}

% \caption{Sample pose estimations by \pmb{$MDN_{3}$} on \textit{coco-val}. An ellipse around each predicted center represents 2d variance terms. Bounding box color indicates mixture mode. More visualizations are provided in the supplementary material.}
% \label{fig:pose_sample}
% \end{figure*}

\vspace{-15pt}
\section{Conclusion and future work}
We propose a mixture formulation for spatial regression in dense 2D object detection and human pose estimation. We show that a mixture density network significantly improves accuracy on both tasks on a real-world large scale dataset. A mixture model provides much better speed-accuracy trade-off and can alleviate the need for multi-scale evaluation. Furthermore, it leads to faster convergence. For both object detection and human pose estimation, a mixture model splits data into meaningful modes; based on object scale and viewpoint, respectively. The models learn to choose the proper output head conditioned on input. 
% We demonstrated that MDNs can be deployed for conditional density estimation on large scale real-world data . 

% We have made it clear that one can use a fully standalone multi-hypothesis model in a real-world scenario without the need to rely on an oracle or postponing model selection to a downstream task. 

In human pose estimation, surprisingly, viewpoint is the dominant factor, and not the pose variation. This stresses that real-world data is multi-modal, but not necessarily in the way we expect. This also motivates using a more sophisticated pose representation in a single-mode model. Designing networks that are able to learn more diverse components is an exciting direction for further research.

% MDN recovers a few modes in this work; however, this also reminds us of the sparsity of latent representations in generative models \cite{xu2019unsupervised}. Designing networks with more diverse components is an exciting direction for future research.

%  Without a principled approach like mixture models, it is difficult to determine the most dominant factors in a data distribution.

% We attribute this to the fact that deep models, even without advanced prediction mechanisms, are powerful enough to deliver relatively high-quality results on the current datasets. 
% \vspace{}
Unlike most works on mixture models, here we use a very diverse large dataset without facing mode collapse. In the future, it will be valuable if one could provide an in-depth study of the role of size and diversity of data in the proper training of mixture models.  Furthermore, it would be insight-full to build a density estimation model on more challenging tasks like the recent large vocabulary instance segmentation task (LVIS) \cite{gupta2019lvis}, which has more than 1000 categories with huge data imbalance. Can mixture models learn to deal with fine modalities on such a diverse dataset?

\vspace{-10pt}
\subsubsection*{Acknowledgments}
This work was partially funded by IMEC through the ICON Lecture+ project and FWO SBO project HAPPY. The computational resources were partially provided by the Flemish Supercomputer Center (VSC).

% \clearpage

{\small
\bibliographystyle{ieee_fullname}
\bibliography{egbib}
}

\end{document}